\documentclass[a4paper,11pt]{article}

\usepackage[english]{babel}
\usepackage[utf8]{inputenc}
\usepackage{amsmath}
\usepackage{amssymb}
\usepackage{amsthm}
\usepackage{bm}
\usepackage{bbm}
\usepackage{cancel}
\usepackage{centernot}
\usepackage{enumerate}
\usepackage{forest}
\usepackage{graphicx}
\usepackage{hyperref}
\usepackage[capitalize,noabbrev]{cleveref}
\usepackage{listings}
\usepackage{mathrsfs}
\usepackage{natbib}
\usepackage{tikz}
\usetikzlibrary{matrix,arrows.meta}
\usepackage[margin=1.0in]{geometry}
\usepackage{algorithm}
\usepackage{algorithmic}

\theoremstyle{plain}
\newtheorem{theorem}{Theorem}[section]
\newtheorem{proposition}[theorem]{Proposition}
\newtheorem{lemma}[theorem]{Lemma}
\newtheorem{corollary}[theorem]{Corollary}
\theoremstyle{definition}

\newtheorem{example}[theorem]{Example}
\theoremstyle{remark}
\newtheorem{remark}[theorem]{Remark}

\newcommand{\bO}{\ensuremath{\mathcal{O}}}
\newcommand{\Pp}{\ensuremath{\mathbb{P}}}

\newcommand{\R}{\ensuremath{\mathbb{R}}}
\newcommand{\Z}{\ensuremath{\mathbb{Z}}}

\newcommand{\N}{\ensuremath{\mathbb{N}}}

\newcommand{\eps}{\epsilon}

\newcommand{\cA}{\ensuremath{\mathcal{A}}}
\newcommand{\cB}{\ensuremath{\mathcal{B}}}

\newcommand{\cH}{\ensuremath{\mathcal{H}}}
\newcommand{\cR}{\ensuremath{\mathcal{R}}}

\newcommand{\B}{\ensuremath{\mathbb{B}}}
\renewcommand{\S}{\ensuremath{\mathbb{S}}}

\newcommand{\hupper}{\overline{h}}

\DeclareMathOperator{\ReLU}{ReLU}

\DeclareMathOperator*{\argmin}{arg\,min}
\DeclareMathOperator*{\argmax}{arg\,max}

\title{Online Learning in Stackelberg Games\\with an Omniscient Follower}
\author{
Geng Zhao \!\thanks{University of California, Berkeley. Email: \texttt{\{gengzhao,banghua\}@berkeley.edu,jiantao@eecs.berkeley.edu,\newline jordan@cs.berkeley.edu}.}\enspace\thanks{The two authors contributed equally to this work.}
\and
Banghua Zhu \!\footnotemark[1]\enspace\footnotemark[2]
\and
Jiantao Jiao \!\footnotemark[1]
\and
Michael I. Jordan \!\footnotemark[1]
}

\date{\today}

\begin{document}

\maketitle

\begin{abstract}
We study the problem of online learning in a two-player decentralized cooperative Stackelberg game. In each round, the leader first takes an action, followed by the follower who takes their action after observing the leader's move. The goal of the leader is to learn to minimize the cumulative regret based on the history of interactions.
Differing from the traditional formulation of repeated Stackelberg games, we assume the follower is omniscient, with full knowledge of the true reward, and that they always best-respond to the leader's actions.  
We analyze the sample complexity of regret minimization in this repeated Stackelberg game. We show that depending on the reward structure, the existence of the omniscient follower may change the sample complexity drastically, from constant to exponential, even for linear cooperative Stackelberg games. This poses unique challenges for the learning process of the leader and the subsequent regret analysis. 
\end{abstract}

\section{Introduction}

The multi-agent learning problem~\citep{ferber1999multi, wooldridge2009introduction, filar2012competitive, zhang2021multi}  has received significant attention reflecting its wide variety of real-world applications, including   autonomous driving~\citep{shalev2016safe, sallab2017deep} and human-robot interaction~\citep{kober2013reinforcement, lillicrap2015continuous, goodrich2008human, xie2021learning}. In a multi-agent system, it is natural to assume that each agent possesses  a different set of information due to its different viewpoint and history of actions. This phenomenon is commonly referred to as the property of \emph{information asymmetry}~\citep{yang2022stackelberg}. %
Such information asymmetry poses challenges to the coordination and cooperation between learning agents.
In this paper, we study how the information asymmetry affects the sample complexity of learning a  two-player decentralized cooperative repeated Stackelberg game, with a focus on the setting when the follower is omniscient and myopic, %
and always best-responds to the leader's actions.

Consider an illustrative example in   human-robot interaction where 
a robot is required to collaborate with a human to achieve some shared objective. This can be formulated as a repeated Stackelberg game where the interactions between human and robot happen in multiple rounds, and the human is an omniscient expert who knows the exact target and how to achieve it. In each round, the robot, as  the leader who  hopes to learn the world model and human behavior from scratch, first takes  some action. After seeing the robot's action, the human, as an expert follower who possesses perfect information about the world, always best-responds to the robot's action to maximize their reward. The robot hopes to use as few as possible interactions to learn the world model and human behavior, and eventually find the optimal action that maximizes a  shared reward.

Concretely, during each round $t$ of the interaction, the leader first plays an action $a_t\in\mathcal{A}$, and the follower plays another action $b_t\in\mathcal{B}$ upon (perfectly) observing $a_t$.  We assume that the two players share a reward, $r_t=h^\star(a_t, b_t)+z_t$, where $z_t\in\mathbb{R}$ is some zero-mean sub-Gaussian noise, $h^\star$ belongs to a family $\mathcal{H}$. We also assume that the follower has full knowledge of the reward and always best responds with $b_t\in\argmax_{b\in\mathcal{B}} h^\star(a_t, b)$. However, the leader does not know $h^\star$ and can only explore via taking actions $a_t$ and making inferences from past observations $(a_1, b_{1}, r_1),\cdots, (a_{t-1}, b_{t-1}, r_{t-1})$.\footnote{For simplicity, we assume in the introduction that the leader can see  $b_1,\cdots, b_{t-1}$ without noise. Later we generalize to the case when the observed $b_t$ is also noisy.}
We are interested in providing tight bound for the Stackelberg regret, defined as
\begin{align*}
    \mathcal{R}(T) = \max_{a\in\mathcal{A}} \mathbb{E}\left[ \sum_{t=1}^T \left(\max_{b\in\mathcal{B}} h^\star(a, b) - \max_{b_t\in\mathcal{B}} h^\star(a_t, b_t) \right)\right].
\end{align*}
The Stackelberg regret characterizes the gap between  the reward achieved from the optimal leader action and the reward from the actual leader action $a_t$.

Compared with the traditional bandit problem, the extra observation of $b_t$ can be viewed as side information accompanying the  usual action-reward pair. Depending on how the function family $\mathcal{H}$ and side information $b$ are designed, 
the complexity of learning for the leader may vary. Here we briefly summarize several illustrative examples where the follower may help or harm the leader's learning process.  We will present a general formalization that encompasses these examples in the next section.

\begin{enumerate}
    \item \textbf{Curse of expertise.} Imagine that in a driving system, the self-driving vehicle (leader) and the human driver (follower)   work together to avoid collisions.  For most of the aggressive actions the leader takes, the final reward for non-collision is high since the human driver will consistently exert efforts to evade the self-driving vehicle in order to prevent collisions. From the leader's point of view, aggressive actions lead to similar outcomes as safe actions. The expertise of the human  prevents the leader from learning from failure cases.
    \item {\textbf{Imitation Learning.}}  %
    Consider an assembly robot (leader)  that learns to move goods to a destination  with a human expert (follower). This can be modeled by the robot choosing a drop-off location, from which the human expert continues to the correct destination. In this simple example, the robot and the human expert cooperate in a ``linear'' fashion---the expert can complete whatever the robot leaves undone, and upon observation of the expert's move the robot should simply imitate the behavior of the human expert  in the future. This corresponds to an ``imitation-based'' interaction that can greatly accelerate the learning process.
\item \textbf{Expert-guided learning.} In most cases, the self-driving vehicle may have some target that is similar but not exactly the same as the human driver. For example, they both aim to avoid collision while heading to a different target. In this case, a pure imitation-based learning will fail. But the self-driving vehicle can still glean good driving standards from the human driver. With the extra observation of the behavior of human driver, the self-driving vehicle can learn much faster.

\end{enumerate}

Extending beyond robotics applications, our framework is potentially applicable in various repeated cooperative game settings where direct communication is hard, unreliable, or forbidden. For instance, it captures the learning aspect of language models adjusting to human preferences, personalized digital healthcare, or other settings of AI-human interaction where explicit revelation of the utility function is difficult: in such settings, the AI system (e.g., the language model, or the digital ``doctor'') works with a human user to achieve a common goal without direct communication of the true preferences or needs; instead, the system must learn them through repeated interactions with the users.

In this paper, we abstract and formalize these three scenarios into a simple linear Stackelberg game and analyze the sample complexity of this game. We briefly overview our main results in the next section. 

\subsection{Main results}
Contrary to the traditional literature on linear bandits, we show that the worst-case sample complexity for achieving $\epsilon$-Stackelberg regret is at least exponential even when $h^\star$ belongs to the linear family $\mathcal{H}_{\phi}=\{\theta\cdot\phi(a, b)\}$. The hard instance corresponds to the `curse of expertise' example discussed above, where the follower's best response hurts the observation, and thus harms the whole learning process. 
\begin{theorem}[Curse of expertise, informal]
  There exists some $\phi$ such that for any algorithm,  we can find some $h^\star\in\mathcal{H}_\phi$ with the regret being $\Omega(T^{(d-3)/(d-2)})$.
\end{theorem}
This shows that the leader needs an exponential number of samples to learn a good policy even when the reward is linear. We also present an upper bound $\bO(T^{(d+1)/(d+2)})$ for linear rewards in Theorem~\ref{Thm_general_exp_upper_bound}.

On the other hand, the side information $b_t$  can also greatly improve the sample complexity when the linear family is structured.   We  provide an  Upper Confidence Bound (UCB) based algorithm~\citep{auer2002finite} that leads to an improved bound in this setting. In particular, we recover the rate for imitation learning when the leader can simply mimic the behavior of the follower.
\begin{theorem}[Imitation learning, informal]
    There exists some $\phi$ such that for any $h^\star\in\mathcal{H}_\phi$,  when $b_t$ is observed, the leader can achieve regret $\bO(\log^ 2(T))$ by imitating the follower behavior. However, when $b_t$ is not observed, the regret is $\Theta(\sqrt{T})$.  
\end{theorem}
Similarly, we can also design cases where observing $b_t$ helps reduce the problem to a traditional linear bandit, while not observing $b_t$ suffers from exponential sample complexity.
\begin{theorem}[Expert-guided, informal]
 There exists some $\phi$ such that for any $h^\star\in\mathcal{H}_\phi$, when $b_t$ is observed, the leader can achieve regret $\bO(\sqrt{T})$. However, when $b_t$ is not observed, the regret is $\Omega(T^{(d-4)/(d-2)})$.
\end{theorem}

In addition to these three examples, we  discuss more complicated scenarios where UCB fails and we show that a careful analysis is necessary to achieve a near-optimal rate. In particular, we establish such a rate for polynomial bandits, where the best-response corresponds to a lower degree polynomial, which helps improve the rate when the noise level for reward and the observed follower behavior is similar. 
\begin{theorem}[Polynomial bandit, informal]
There exists a family of  $2k$-degree polynomial, such that the regret is  $\Theta(\sqrt{d^{2k-1}T})$ when $b_t$ is observed, and $\Theta(\sqrt{d^{2k}T})$ when $b_t$ is not observed.  
\end{theorem}

\subsection{Related work}

\paragraph{Decentralized  Stackelberg Games.}
The problem of repeated Stackelberg games has been studied extensively~\citep{von2010market, marecki2012playing, lauffer2022no, kao2022decentralized}, in a standard setting where the leader leads and the myopic follower follows with its best response for the current round.

\citet{kao2022decentralized}  and \citet{lauffer2022no} study a similar setting to ours, in which  a leader and a follower interact through a cooperative Stackelberg game that comprises a two-stage bandit problem.
However, \citet{kao2022decentralized} restrict their focus to   the tabular case where both $\cA$ and $\cB$ are finite and the reward $h^\star$ is uncorrelated for different actions $(a, b)$. They also assume that both the leader and the agent are running regret-minimization algorithms independently.   They show that the classic upper confidence bound (UCB) algorithm for the multi-arm bandit problem can be used for both the leader and the agent, respectively, to achieve asymptotically optimal performance (i.e., no-regret). However, it is unclear that such results can generalize to bandits with function approximation and the case of omniscient agents. Indeed, our results show that the general case (or even just the linear case) is not always statistically tractable. Note also that \citet{lauffer2022no} show that the regret can depend exponentially on the  dimension of the agent's utility.

Other examples of Stackelberg games include Stackelberg security games~\citep{conitzer2006computing, tambe2011security}, strategic learning~\citep{hardt2016strategic, dong2018strategic,liu2016bandit},  dynamic task pricing~\citep{kleinberg2003value} and online contract design~\citep{ho2014adaptive, zhu2022sample}. 
The problem of online learning in contract theory considers a decentralized general-sum Stackelberg game with omniscient agents. It focuses on a special case where the rewards for the leader and the agent are both linear. It is shown in~\citet{zhu2022sample} that one has to pay exponential sample complexity in this setting to achieve small regret in the worst case. 

\paragraph{Centralized  Stackelberg Game.}
Centralized Stackelberg games are also well studied in the literature~\citep{zhong2021can, bai2021sample,gerstgrasser2022oracles, yu2022learning}, where the machine learning algorithm has control over both the leader and the follower.
\citet{bai2021sample} consider the repeated Stackelberg game where both the leader and the agent learn their optimal actions (a Stackelberg equilibrium) from samples. However, they assume a central controller that can determine the actions of both the leader and the agent. Moreover, they rely on an assumption of a bounded gap between the optimal response and an $\epsilon$-approximate best response. %
In contrast, in our framework, we assume that the agent's utility is unknown, and that the agent always takes the best response. 

\paragraph{Bandit with side information.}
There has been significant effort in studying bandits with side information~\citep{wang2005bandit, langford2007epoch, foster2021statistical}. Such side information is generally assumed to be available before a decision. \citet{foster2021statistical} also consider the case when an extra observation is available after taking the actions. However, they mainly focus on the setting of reinforcement learning where the extra observation is the trajectory. Although our observation of follower behavior can also be viewed as side information, it also alters the reward in the  Stackelberg game, which changes the structure of the multi-agent problem.

\section{Formulation}\label{Sec_formulation}

We consider a two-player cooperative Stackelberg bandit game with an omniscient follower.

Let $\cA\subseteq \R^{d_1}$ and $\cB\subseteq \R^{d_2}$ be compact sets.  Up to a scaling factor, we will assume that $\cA$ and $\cB$ reside inside the unit ball centered at the origin. During each round $t\in[T]$ of interaction, the leader plays an action $a_t \in \cA$, and the follower plays $b_t\in\cB$ upon (perfectly) observing $a_t$. The two players both receive a reward $r_t = h^\star(a_t,b_t) + z_t$, where $z_t\in\R$ is zero-mean $\sigma_r$-sub-Gaussian and is independent of all past events. We will make the realizability assumption that $h^\star$ belongs to a (known) family $\cH$ of real-valued functions on $\B^{d_1} \times\B^{d_2}$. As is common in the study of bandits, we assume that reward function is bounded, i.e., there exists $C\in(0,\infty)$ such that $0 \le h \le C$ for all $h\in\cH$. We assume $C=1$ throughout the paper unless  stated otherwise. %

We will assume that the follower, modeled after an expert human player, has full knowledge of the game and can always best respond with an optimal action $b_t \in \argmax_{b\in\cB} h^\star(a_t,b)$. The leader then makes a noisy observation of $b_t$, given by $\hat{b}_t = b_t + w_t$, where $w_t \in\R^{d_2}$ is zero-mean $\sigma_b$-sub-Gaussian (e.g., component-wise $\sigma_b$-sub-Gaussian with independent zero-mean coordinates) and independent of all past events.

For convenience, we denote the set of best responses to leader's action $a$ when the ground truth reward function is $h$ by $b^*_h(a)$. Denote $\hupper(a) := \max_{b\in\cB} h(a,b)$. The optimal action, unbeknownst to the leader, is denoted $a^* := \argmax_{a\in\cA} \hupper^\star(a)$.

The leader's objective is to minimize the \emph{regret} during $T$ rounds of interactions, defined as
\begin{equation}
    \cR(T) = \max_{a\in\cA} \mathbb{E}\left[\sum_{t=1}^T \hupper(a) - \hupper(a_t)\right].
\end{equation}
We will also focus on the sample complexity of achieving low (average) regret; that is, for some $\eps,\delta\in[0,1]$, the minimal $T\in\N$ such that $\cR(T) \le \eps T$. %

\paragraph{Notations.} 
We use calligraphic letters for sets and operators, e.g., $\mathcal{A}$. Given a set $\mathcal{A}$, we write $|\mathcal{A}|$ for the cardinality of $\mathcal{A}$. $\B^d$ and $\S^{d-1}$ denote the unit ball and the unit sphere, both centered at the origin, in $d$-dimensional Euclidean space. Vectors are assumed to be column vectors except for the probability and measure vectors. %
For a vector $v\in\mathbb{R}^d$ and an integer $i\in\mathbb{N}$, 
we use $v_i$ to denote the $i$-th element of $v$, and $v_{-i}$ to denote the vector of all elements in $v$ except for $v_i$. 
For two $n$-dimensional vectors $x$ and $y$, we use $x \cdot y = x^\top y$ to denote
their inner product.  We write $f(x) = \bO(g(x))$ or $f(x)\lesssim g(x)$ if   there exists some positive real number $M$ and some $x_0$   such that $|f(x)|\leq M g(x)$ for all $x\geq x_0$.  We use $\widetilde \bO(\cdot)$ to be the big-$\bO$ notation ignoring logarithmic factors. We write $f(x) = \Omega(g(x))$ or $f(x)\gtrsim g(x)$ if   there exists some positive real number $M$ and some $x_0$   such that $|f(x)|\geq  M g(x)$ for all $x\geq x_0$.
We write $f(x) = \Theta(g(x))$ if we have both $f(x) = \bO(g(x))$  and $f(x) = \Omega(g(x))$. We use    
$\|\cdot\|_p$ to denote the $\ell^p$ norm for $p\in(0,\infty]$, with $\|\cdot\|$ denoting the Euclidean ($\ell^2$) norm $\|\cdot\|_2$. %

\paragraph{Parameterized family.} In subsequent discussions, we will consider the parameterized case when $\cH$ admits a parameterization over a compact parameter space $\Theta$. The class is denoted by $\cH_\Theta = \{h_\theta | \theta\in\Theta\}$. When the parameterization is linear, that is,
\begin{equation}
    h_\theta(a,b) = \theta \cdot \phi(a,b)
\end{equation}
for some feature function $\phi:\cA\times\cB\to\B^d$, we will denote the class by $\cH_{\Theta,\phi}$. We denote the true parameter by $\theta^\star$.
For instance, when $\cA$ and $\cB$ are the sets of standard basis vectors in $\R^{|\cA|}$ and $\R^{|\cB|}$ with $\phi(a,b)=ab^\top$ and $\theta$ is bounded in $\R^{|\cA|\times|\cB|}$, we recover the tabular case model in \citet{kao2022decentralized} with finite action sets. In general, however, we will focus on cases with infinite action sets.

\section{Linear Stackelberg games: Curse of expertise}\label{Sec_basic_upper_lower_bounds}

In this section, we study the sample complexity of learning in linear Stackelberg game, where the family of reward is restricted to $\mathcal{H}_{\Theta,\phi}$ for some given $\Theta$ and $\phi$.

\subsection{An exponential lower bound}

It is well known that the regret for traditional linear bandits grows as $ \Theta(d\sqrt{T})$~\citep{abbasi2011improved}. In the case of a linear Stackelberg game,  
we   present a worst-case lower bound on the regret  that is exponential in dimensionality for the linear family. This suggests that the leader cannot learn the task well unless in possession of an exponential number of samples even when we restrict to linear Stackelberg games. 

Assume the leader makes perfect observations of the follower's responses (i.e., $\sigma_b = 0$). We have the following  lower bound.

\begin{theorem}\label{Thm_exp_lower_bound}
For any $d\geq 4$,
there exists some $\phi$ such that, for any algorithm that the leader runs, one can find some   instance with $h_\theta\in\cH_{\Theta,\phi}$ such that
\begin{equation}
   \cR(T) \gtrsim T^{{(d-4)}/{(d-2)}}.
\end{equation}
\end{theorem}
In other words, the  sample complexity for achieving $\eps$ (average) regret is at least $\Omega\big((1/\epsilon)^{\frac{d-2}{2}}\big)$.

The proof is detailed in Appendix~\ref{proof:Thm_exp_lower_bound}.
The worst-case instance presented below can be reduced to the ReLU bandit problem shown below, which is known to suffer from the exponential sample complexity \citep{dong2021provable}. 

\begin{example}\label{ex:relu_bandit}
    Let $\cA = \B^{d-1}$, $\cB = [0,1]$ and $\Theta = \{\theta \mid \theta_{-d}\in \S^{d-2}, \theta_d = 1-\Delta\}$ for some $\Delta\in (0,1)$. Let the feature function be $\phi(a,b) = ((1-b)a, b)$.
\end{example}

One can verify that in this case, one has
\begin{equation}
    \hupper_\theta(a) = \max\{1-\Delta, \theta_{-d}\cdot a\}.
\end{equation}
Thus when $a$ is chosen far from $\theta_{-d}$, the reward will remain constant.

Theorem~\ref{Thm_exp_lower_bound} is no mystery mathematically: the best response may destroy linearity for the leader's observations, imposing a toll. Conceptually, however, the message from the theorem is striking: it highlights a ``curse of expertise''; i.e., the potential difficulty to learn with an expert on a decentralized bandit learning task with a large action space.
From the classic single-agent bandit learning perspective, the task the two agents aim to solve is straightforward: a linear bandit on an action space $\phi(\cA,\cB)$. In other words, if the expert follower lets the novice leader control the choice of $b$, the average regret would steadily decrease at a rate of $\widetilde{\bO}(d\sqrt{T})$.  On the other hand, with a myopic focus, the follower's expertise in best responding ironically results in a significantly higher regret, as it deprives the learner of the ability to explore.

In the context of autonomous driving, for example, this can manifest in scenarios where the autonomous vehicle takes a poor action (e.g., an aggressive lane change) yet other vehicles or pedestrian immediately respond by slowing down or steering away to avoid a possible collision, thereby hiding the potential negative consequences of the action. The lack of coordination and the constant best response from the follower, both common in practice, makes it hard for the leader to efficiently learn the reward landscape or improve their current policy. %

\subsection{An exponential upper bound}

For any class $\cH$ of reward functions on a pair of actions $(a,b)$, an upper bound on the sample complexity (and regret) can be obtained using a covering argument.

\begin{theorem}\label{Thm_general_exp_upper_bound}
Let $N(\eps) = N(\cH,\eps,\|\cdot\|_\infty)$ denote the $\ell^\infty$ covering number of $\cH$ with radius $\eps > 0$. Then we can achieve
\begin{equation}
    \cR(T) \lesssim \inf_{\eps > 0}\eps T + \sqrt{N(\eps)T}.
\end{equation}
\end{theorem}

To achieve this, simply compute an $\eps$-covering of $\cH$ and let the leader play no-regret algorithms on  the $\eps$-covering set. Note that although the covering is constructed for pair of actions $(a, b)\in\mathcal{A}_\epsilon\times \mathcal{B}_\epsilon$, it suffices for the leader to run no-regret algorithms on actions $\mathcal{A}_\epsilon$. The detailed algorithm and proof are given in Appendix~\ref{proof:Thm_general_exp_upper_bound}.

This upper bound is achieved when the leader does not even utilize the observations of the follower's responses. Indeed, in the worst case (e.g., in Example ~\ref{ex:relu_bandit}), the responses will not provide  information.

As a corollary, in the linear regime with $\cH_{\Theta,\phi}$,
the covering number is $N(\eps) = N(\Theta,\eps,\|\cdot\|) \le \exp\big(O\big(d\log\frac{1}{\eps}\big)\big)$ \citep{wainwright2019high}. Choosing $\eps \asymp T^{-{1}/{(d+2)}}$, Theorem~\ref{Thm_general_exp_upper_bound} reduces to the following upper bound in the linearly parameterized case.

\begin{corollary}\label{cor:linear}
    In the linear case, we can achieve $\cR(T) \lesssim T^{{(d+1)}/{(d+2)}}$.
\end{corollary}

In other words, the sample complexity for achieving average regret equal to $\eps$ is upper bounded by $\bO\big(\big({1}/{\eps}\big)^{d+2}\big)$.
 This upper bound is agnostic to any structural property of the feature function $\phi$, such as smoothness or even continuity.

\section{UCB with side observations}\label{Sec_ucb}
Although the worst-case sample complexity for linear Stackelberg games is exponential, it is possible to obtain a fine-grained analysis and improved rate for the family $\mathcal{H}_{\Theta,\phi}$ when  $\phi$ is better structured. 
A natural choice of algorithm for the leader is some variant of UCB that incorporates observations of the follower's actions.
In this section, we will describe a general recipe for a family of UCB algorithms to incorporate the side information as well as the challenge in their design.

\subsection{Algorithm description}

We consider the following variant of UCB that uses the follower's responses as side information to improve the confidence set.

\begin{algorithm}[hb]
   \caption{UCB with side information from expert}
   \label{alg:ucb}
\begin{algorithmic}
   \STATE {\bfseries Input:} Regression oracles $\mathsf{Reg}^{(b)}$ and $\mathsf{Reg}^{(r)}$ on reward and response, $\{\alpha_t\}_{t\in[T]}$, $\{\beta_t\}_{t\in[T]}$
   \FOR{$t=1$ {\bfseries to} $T$}
   \STATE Compute $h_t^{(b)}=\mathsf{Reg}^{(b)}(\hat{b}_1,\ldots,\hat{b}_{t-1})$ and $h_t^{(r)}=\mathsf{Reg}^{(r)}(r_1,\ldots,r_{t-1})$
   \STATE Set $\cH_t^{(b)} := \{h: \sum_{i=1}^{t-1} \|b^*_h(a_i) - b^*_{h_t^{(b)}}(a_i)\|^2 \le \alpha_t^2\}$
   \STATE Set $\cH_t^{(r)} := \{h: \sum_{i=1}^{t-1} \big(\hupper(a_i) - \hupper_t^{(r)}(a_i)\big)^2 \le \beta_t^2\}$
   \STATE Construct confidence set $\cH_t = \cH_t^{(b)} \cap \cH_t^{(r)}$
   \STATE Take action $a_t \in \argmax_{a\in\cA} \sup_{h\in \cH_t} \hupper(a)$
   \STATE Observe (noisy) reward $r_t$ and response $\hat{b}_t$
   \ENDFOR
\end{algorithmic}
\end{algorithm}

\begin{remark}
    The regression oracles and the sequences $\{\alpha_t\}_{t\in[T]}, \{\beta_t\}_{t\in[T]}$ must be chosen appropriately so that the following condition holds:
Given an error tolerance $\delta \in (0,1)$, we require $h^\star\in\bigcap_{t=1}^T \cH_t$ with probability at least $1-\delta$. 
\end{remark}

\begin{remark}
A common choice for $\mathsf{Reg}^{(b)}$ and $\mathsf{Reg}^{(r)}$ is the least-squares regression oracle that computes
\begin{equation}\label{eqn_least_sq_1}
    h_t^{(b)} \in \argmin_{h\in\cH} \sum_{i=1}^{t-1} \|b^*_h(a_i) - \hat{b}_i\|^2
\end{equation}
and
\begin{equation}\label{eqn_least_sq_2}
    h_t^{(r)} \in \argmin_{h\in\cH} \sum_{i=1}^{t-1} (\hupper(a_i) - r_i)^2.
\end{equation}
When the least-squares computation becomes infeasible under complex response-reward structures (this is common for \eqref{eqn_least_sq_1}), custom oracles need to be designed. A more intricate approach may be to jointly construct the estimate using both $\{\hat{b}_\tau\}_{\tau\in[t-1]}$ and  $\{r_\tau\}_{\tau\in[t-1]}$. We leave it for future research to study systematic designs of the oracles and the confidence sets.
\end{remark}

\begin{remark}
When the responses are unobserved or ignored (e.g., by choosing $\alpha_t = \infty$), Algorithm~\ref{alg:ucb} reduces to the classic Eluder UCB using the least-squares (reward) oracle 
with $\cH_t = \cH_t^{(r)}$~\citep{russo2013eluder}.
\end{remark}

The choices of $\{\alpha_t\}_{t\in\N}$ and $\{\beta_t\}_{t\in\N}$ can pose another challenge. An naive attempt to get a generic upper bound on $\alpha_t$ is to use a covering argument as in \citet{russo2013eluder} using the following measurement between two functions $h,h'\in\cH$: $d^{(b)}(h,h') = \sup_a \|b^*_h(a) - b^*_{h'}(a)\|$. But note that this does not necessarily define a norm, and further the covering number of $\cH$ in this sense can be infinite when the best response is discontinuous in the leader's action $a$. Thus, such an approach is often not useful and one may have to determine $\alpha_t$ on a per instance basis.

\subsection{Examples}
While Theorem~\ref{Thm_exp_lower_bound} shows that the involvement of the omniscient follower can lead to ``curse of expertise,'' a stark deterioration in the sample complexity, there are many scenarios where the leader's observation of the follower's responses can expedite learning significantly. In this section, we will explore a few such examples.

\subsubsection{An imitation-based example}\label{Sec_example_1}

Let us consider a setting where the leader achieves efficient learning through imitation. Heuristically, imitation arises when the optimal action for the leader is equal to the best response for the omniscient follower or a function of it. This may capture, for instance, real-world robotics applications where the actions of the robot and the human expert are exchangeable and the true goal can be easily inferred from the expert's action. A simple scenario is when the robot and the human expert are supposed to carry out the same task perfectly, in which case the robot should simply treat the expert as a role model and imitate. The following is a concrete example.

\begin{example}\label{Ex_theta_aplusb}
    Let $\cA = \cB = \Theta =\S^{d-1}$ (or $\B^d$ equivalently)\footnote{While it is customary to consider $\Theta=\B^d$, we will observe below that the imitation-based algorithm does not crucially rely on $\|\theta^\star\|$ and only incurs smaller regret if $\|\theta^\star\|<1$. This is because the algorithm asymptotically relies solely on the response observations, which are invariant under scaling of $\theta^\star$. It is also without loss of generality to restrict all actions to the sphere.}. Consider the linearly parameterized function class $\cH_{\Theta,\phi}$ with feature function
    \begin{equation}
        \phi(a,b) = a + b.
    \end{equation}
\end{example}

Here, the optimal response $b^*_\theta \equiv \theta$ is independent of $a$, and $\hupper_\theta(a) = \theta\cdot a + 1$.

\paragraph{Construction of confidence sets.}
The (noisy) observations of the follower's best responses simplify the problem into an imitation learning task.
A simple oracle for the best-response observations is to take the $\cA$-projected empirical average of responses, i.e., $\theta_t^{(b)} = \Pi_{\cA}\big(\frac{1}{t-1}\sum_{i=1}^{t-1}\hat{b}_i\big)$.\footnote{Define the projection of $y\in\R^d$ onto a closed set $\mathcal{X}\subseteq\R^d$ as $\Pi_\mathcal{X}(y) := \argmin_{x\in\mathcal{X}} \|y-x\|$, breaking ties arbitrarily when the minimizer is not unique.}
 The response-based confidence set reduces to
\begin{equation*}
    \Theta_t^{(b)} = \Big\{\theta\in\Theta \Big| \|\theta - \theta_t^{(b)}\| \le \frac{\alpha_t}{\sqrt{t-1}}\Big\}.
\end{equation*}
Standard sub-Gaussian concentration results suggest that the (Euclidean) radius of this confidence set shrinks at a rate of $t^{-1/2}$.
\begin{lemma}\label{Lemma_choice_of_alpha}
    To ensure $\theta^\star \in \bigcap_{t\in [T]} \Theta_t$ with probability at least $1-\delta$, it suffices to choose $\alpha_t = \Theta\big(\sigma_b\sqrt{ d +\log \frac{T}{\delta}}\big)$.
\end{lemma}

UCB chooses actions on $\S^{d-1}$ increasingly close to the empirical estimate $\theta_t^{(b)}$.\footnote{Even simpler, the leader can play the $\cA$-projected empirical average of responses. Under our choice of constant $\alpha$, the analysis will be the same, with the result differ by at most a constant factor.}
The regret bound follows from these choices of confidence sets.

\begin{proposition}\label{Prop_theta_a_b_upper_bound}
    In Example~\ref{Ex_theta_aplusb}, UCB achieves a regret bound
    \begin{equation}
        \cR_{UCB}(T) \lesssim \sigma_b^2 \log T \cdot(d +\log T).
    \end{equation}
\end{proposition}
In other words, the average regret decays at a rate of $\widetilde{\bO}(\sigma_b^2 d / T)$. This has also been analyzed in the setting of imitation learning~\citep{rajaraman2021value}, and the results are consistent.

\begin{remark}
    When the follower's responses are unobserved (still assumed to be best responses), this is simply a linear bandit, where the minimax regret is $\Omega(\sigma_b d\sqrt{T}) \gg \bO(\sigma_b^2 d \log^2 T)$. This indicates the value of the $b_t$ observations. When the follower's response is noiseless, one can see that a single sample suffices to find the optimal response since one always observes $b^\star_\theta =\theta$.
\end{remark}

\begin{remark}
    Note the gap in the $\Theta(\log T)$ regret when the response observations are used and the $\Theta(\sqrt{T})$ regret when they are ignored or unavailable, showing the value of those response observations. In fact, it is easy to modify this example slightly (e.g., taking $\phi(a,b) = \max\{|\theta^\top a|, \Delta\} b$ for some $\Delta\in(0,1)$) to create an even larger gap: When the leader uses the response observations, the regret is $\widetilde{\bO}(d\log T)$ with sample complexity $\widetilde{\bO}\big(d\log\frac{1}{\eps}\big)$; When the response observations are unavailable, the sample complexity increases to $\Omega(\eps^{-d})$.
\end{remark}

\subsubsection{Expert-guided exploration}

In many scenarios, the omniscient follower's actions may not directly reveal the exact state of the world but still provide crucial information. The next example illustrates a simple setting where the follower's response can significantly reduce the sample complexity.

\begin{example}\label{Ex_guide_explore}
    Let $\cA = \cB = \S^{d-1}$ and
    \begin{equation*}
        \Theta = \{(\theta_a,\theta_b) \in \S^{d-1}\times\S^{d-1} | \theta_a\cdot \theta_b \ge \zeta\}
    \end{equation*}
    for some $\zeta\in(0,1)$. Consider the parameterized family of functions $\cH_\Theta = \{h_\theta | \theta\in\Theta\}$ where
    \begin{equation*}
        h_\theta(a,b) = \ReLU(\theta_a \cdot a - \Delta)+\theta_b\cdot b,
    \end{equation*}
    for some $\Delta \in(0,1)$. For simplicity, we will assume that the response observations are noiseless (i.e., $\sigma_b = 0$), although the noisy case can be analyzed analogously.
\end{example}

\paragraph{Confidences sets.} The best response is $b^*_{\theta} \equiv \theta_b$, again independent of the leader's action.  Upon observing $b_1 = \theta_b$, the leader should construct confidence sets $\Theta_t^{(b)} = \{\theta_a\in\S^{d-1}|\theta_a\cdot b_1\ge \zeta\}\times\{b_1\}$, while $\Theta_t^{(r)}$ is chosen as in linear UCB. As a result, all subsequent actions the leader takes must fall into
\begin{equation}
    \cA_1 := \{a\in\cA | a\cdot b_1 \ge \zeta\}.
\end{equation}
This refinement of the action set will reduce the sample complexity, and depending on the size of $\zeta$ relative to $\Delta$, the reduction can be significant.

\paragraph{Strong reduction.} When $1-\zeta \le (1- \Delta)/4$, the leader learns that $\theta_a \cdot b_1 \ge \zeta$. In particular, any action $a\in\cA_1$ must satisfy
\begin{multline}
    \theta_a \cdot a = \frac{2-\|\theta_a-a\|^2}{2} \ge \frac{2-(\|\theta_a-b_1\|+\|a-b_1\|)^2}{2} \\
    \ge \frac{2-(2\sqrt{2-2\zeta})^2}{2} = 1 - 4(1-\zeta) \ge \Delta,
\end{multline}
and thus $\hupper(a) = \theta_a\cdot a - \Delta + 1$ behaves as a linear function within $\cA_1$. By playing UCB within $\cA_1$, the leader reduces the problem to a linear bandit instance and thus achieves the following regret bound.

\begin{proposition}\label{Prop_guide_explore_upper_bound_strong_reduction}
    Assume $1-\zeta \le (1- \Delta)/4$ in Example~\ref{Ex_guide_explore}. UCB achieves
    \begin{equation}\label{Eqn_guided_explore_strong}
        \cR_{UCB}(T) \le \widetilde{\bO}(d\sqrt{T}).
    \end{equation}
\end{proposition}
This leads to a sample complexity of $\widetilde{\bO}(d^2/\eps^2)$,
in contrast to the exponential sample complexity $\exp(\bO(d\log\frac{1}{\eps}))$ if the responses were unobserved. Information from the follower's response guides the leader's exploration to the well conditioned part of the action space. Given the $\Omega(d\sqrt{T})$ sample complexity of linear bandits, the upper bound \eqref{Eqn_guided_explore_strong} is tight (up to logarithmic terms).

\paragraph{Weak reduction.} When $\zeta$ is small relative to $\Delta$, the problem does not immediately reduce to a linear bandit, but we have the following improved upper bound.

\begin{proposition}\label{Prop_guide_explore_upper_bound_weak_reduction}
    There exists an algorithm $\mathsf{Alg}$ that achieves
    \begin{equation}\label{Eqn_prop_guide_explore_weak_reduction_main}
        \cR_\mathsf{Alg}(T) \le \bO\big((C_\zeta^d T^{d+1})^{\frac{1}{d+2}}\big),
    \end{equation}
    where $C_\zeta := \sqrt{1-\zeta^2} \in (0,1)$.
\end{proposition}
This bound improves as $\zeta$ decreases. The sample complexity is therefore $\widetilde{\bO}(C_\zeta^d \eps^{-d-2})$, a $C_\zeta^d$ reduction compared with the original complexity without observing the responses in Corollary~\ref{cor:linear}.

Since the reduced problem is still a ReLU bandit, UCB will not be suitable. Instead, \eqref{Eqn_prop_guide_explore_weak_reduction_main} can be achieved through discretization of $\cA_1$ as the upper bound in Theorem~\ref{Thm_general_exp_upper_bound}.

\section{Beyond UCB}\label{Sec_beyond_ucb}
Although the UCB algorithm gives a near-optimal rate in most of the above examples. We also provide two cases where UCB fails to achieve the optimal rate. This necessitates a tailored algorithm design in specific settings.
\subsection{Nonlinear (polynomial) family}

UCB is known to fail to achieve the optimal rate in the case of the polynomial bandit family \citep{huang2021optimal}, where the reward is a polynomial activation on top of a linear family. We construct an example which utilizes the structure of the polynomial bandit, formally defined  below.

\begin{example}[Polynomial bandit]\label{Ex_nonlinear}
Consider the convex function $f(x) = x^{2k}$ for some $k\in\Z_+$. Let
\begin{equation}
    \cA = \B^{d-1}, \cB=[-1,1], \Theta = \B^{d-1}\times\{1\},
\end{equation}
and
\begin{equation}
    \phi(a,b) = (2kba, -f^*(2kb)),
\end{equation}
where $f^*$ is the convex conjugate of $f$. Consider the nonlinearly parameterized family
\begin{equation}
    \cH_\Theta:=\{h_\theta(a,b) = f(\theta \cdot \phi(a,b)) \mid \theta\in\Theta\}.
\end{equation}
\end{example}

By properties of the  convex conjugate,
\begin{equation}
    \hupper_\theta(a) = f(\theta_{-d}\cdot a) = (\theta_{-d}\cdot a)^{2k}
\end{equation}
with the best response
\begin{align*}
    b^*_\theta(a) &= \argmax_{-1\le b \le 1} 2kb \theta_{-d}\cdot a - f^*(2kb) \\
    &= \frac{ f'(\theta_{-d} \cdot a)}{2k}  = (\theta_{-d}\cdot a)^{2k-1} \in [-1,1].
\end{align*}
This observation allows us to apply results on polynomial bandits \citep{huang2021optimal}.

\paragraph{Response-regret structure.} Observe the following properties of the best response function in Example~\ref{Ex_nonlinear}.
\begin{enumerate}
    \item\label{Obs_1} The expected reward is a function of the best response, independent of the true parameter. Namely,
    \begin{equation}
        \hupper_\theta(a) = b^*_\theta(a)^{\frac{2k}{2k-1}}.
    \end{equation}
    This mapping is Lipschitz:
    \begin{equation}\label{Eqn_ex_nonlinear_lipschitz}
        \big|\hupper_\theta(a) - \hupper_\theta(a')\big| \le \frac{2k}{2k-1}\big| b^*_\theta(a) - b^*_\theta(a') \big|,
    \end{equation}
    and further
    \begin{equation}
        \argmax_{a\in\cA} b^*_\theta(a) = \theta \in \argmax_{a\in\cA} \hupper_\theta(a),
    \end{equation}
    with both maxima being 1.

    \item\label{Obs_2} The response observation, as a degree $2k-1$  polynomial, is more informative than the reward observation, a degree $2k$ polynomial, when the noise levels are the same and $\theta_{-d}\cdot a$ is small.
\end{enumerate}
Based on these two observations, the leader may view the response $b_t$ as a \emph{proxy reward} and aim to minimize the \emph{proxy regret} 
\begin{equation}
    \widehat{\cR}(T) := \sum_{t=1}^T 1-b^*_\theta(a_t).
\end{equation}
This is consistent with minimizing the true regret $\cR(T)$, which differs from the proxy regret $\widehat\cR(T)$ by at most a constant factor by \eqref{Eqn_ex_nonlinear_lipschitz}.

\paragraph{Regret bound.} Using the response observations exclusively to minimize the proxy regret $\widehat{\cR}(T) = \sum_{t=1}^T 1-b^*_\theta(a_t)$, the leader reduces her task to a polynomial bandit problem with a degree $2k-1$ polynomial activation function.
By \eqref{Eqn_ex_nonlinear_lipschitz}, we may focus on bounding the proxy regret. Corollary~3.16 from \citet{huang2021optimal} suggests that
\begin{equation}\label{Eqn_ex_nonlinear_proxy_regret_bound}
    \widehat{\cR}(T) \le \widetilde{\bO}(\sqrt{d^{2k-1}T}),
\end{equation}
or equivalently the sample complexity is $\widetilde{\bO}(d^{2k-1}/\eps^2)$ for achieving $\eps$ average proxy regret. The following bound on the true regret follows from \eqref{Eqn_ex_nonlinear_lipschitz} and \eqref{Eqn_ex_nonlinear_proxy_regret_bound}.

 \begin{proposition}\label{Prop_nonlinear_bandit_upper}
    In example~\ref{Ex_nonlinear}, there exists an algorithm $\mathsf{Alg}$, using the response observations exclusively, that achieves
    \begin{equation}\label{Eqn_prop_nonlinear}
         \cR_{\mathsf{Alg}}(T) \le \bO(\sqrt{d^{2k-1} T}).
     \end{equation}
 \end{proposition}
 
Proposition~\ref{Prop_nonlinear_bandit_upper} suggests an $\widetilde{\bO}(d^{2k-1}/\eps^2)$ sample complexity. For instance, the leader can achieve this regret with the zeroth-order algorithm proposed in \citet[Algorithm~6]{huang2021optimal}.

\begin{remark}[Lower bound]
    Since the reward observations have a higher signal-to-noise-ratio, we should expect that the sample complexity of Example~\ref{Ex_nonlinear} to be the same order as the sample complexity of achieving $\eps$ average regret in a degree $2k-1$ polynomial bandit. \citet{huang2021optimal} shows that this is lower bounded by $\Omega(d^{2k-1}/\eps^2)$. Thus, \eqref{Eqn_prop_nonlinear} is essentially optimal.
\end{remark}
\begin{remark}
    [Benefit of observing responses] If the leader does not observe the responses, the problem is equivalent to a degree $2k$ polynomial bandit. 
    The optimal regret without observing the experts actions will lead to an $\widetilde{\bO}(d^{2k}/\eps^2)$ sample complexity.
    Thus, the response observations contribute to shaving of a factor of $d$, which can be significant when the dimensionality is high.
\end{remark}

\begin{remark}
    [Suboptimality of UCB] Using the traditional Eluder UCB algorithm leads to a suboptimal sample complexity of $\widetilde{\bO}(d^{2k}/\eps^2)$ when the leader solely uses the response observations. Still, this is a factor $d$ improvement compared to what she can achieve with UCB without the response observations.
\end{remark}

\subsection{Failure of the optimism principle}

The next example is adapted from the ReLU bandit in Example~\ref{ex:relu_bandit}, and shows that optimism-based method can have dramatic suboptimality in certain problems.

\begin{example} 
\label{Ex_fail_ucb}
Let $\cA = \B^{d-1}, \cB = \B^{d-1}\times[0,1]$, and 
\begin{equation}
    \Theta = \{(\theta_{-d},\theta_d)\mid \theta_{-d}\in\B^{d}, \theta_d=1-\Delta\}
\end{equation}
for some $\Delta\in(0,1)$. Consider the linear family $\cH_{\Theta,\phi}$ with
\begin{equation}
    \phi(a,b) = \|a\| ((1-b_d) a, b_d-\|b_{-d}\|) + \frac{1-\|a\|}{2}(b_{-d},0).
\end{equation}
\end{example}

For any $\theta\in\Theta$ with $\theta_{-d}\in\S^{d-1}$, the optimal action for the leader is 
$\theta_{-d}$, with the follower best responding $(0,0)$ and achieving unit expected reward.

When $\|a\| = 1$, this function behaves exactly as in Example~\ref{ex:relu_bandit}, where $b^*_\theta(a) = (0,1)$ whenever $\theta_{-d}\cdot a < 1-\Delta$; When $a=0$, the best response is $b^*_\theta(0) = (\theta_{-d},b_d)$. Thus, if the response observations are noiseless, the leader learns the true parameter and hence the optimal action in one round by playing $a_1 = 0$.

However, any optimism-based method such as UCB will not achieve such efficient learning, even when the response are noiselessly observed.
It is straightforward to verify that, for any action $a$ with $\|a\|< 1$, the optimistic reward satisfies
\begin{equation}
    \sup_{\theta\in\Theta} \hupper_\theta(a) < 1.
\end{equation}
Thus, as long as the confidence set contains some $\theta$ with $\theta_{-d} \in\S^{d-1}$, which holds under our initial condition, optimism causes the leader to only take actions $a\in\S^{d-1}$, reducing the problem to the worst-case Example~\ref{ex:relu_bandit}.

\section{Conclusions}

We have studied a model of online learning in decentralized cooperative Stackelberg games.
We showed that, even with an omniscient follower who always best responds (myopically), the worst case sample complexity for a linear family can be as large as $\exp(\Theta(d\log \frac{1}{\eps}))$.
This ``curse of expertise'' highlights the challenge caused by miscoordinated exploration. This also raises the question of how a non-myopic expert follower should respond to the leader's actions (without knowing the leader's exact algorithm) to expedite their learning and maximize their long-term reward.

We  considered the UCB-type algorithm that incorporates response observations.  A few examples of various hardness were considered, ranging from efficient learning through imitation and guided exploration to the worst-case linear family example with an exponential sample complexity.

Besides the  examples considered in the paper, there are numerous scenarios where  the roles of the leader and the follower are more complex to reason about. This poses unique challenges for both the learning process of the leader and the subsequent analysis of regret, indicating a fertile ground for future research. Specifically, our current template of Algorithm~\ref{alg:ucb} requires designing the confidence sets based on the specific response-reward structure of each problem. It remains open to find a general design (or prove the lack thereof) that systematically synthesizes the response and reward observations. A general framework of analysis that can provide a unified yet sharp upper bound on the examples is also valuable.

\bibliography{ref}

\begin{thebibliography}{37}
\providecommand{\natexlab}[1]{#1}
\providecommand{\url}[1]{\texttt{#1}}
\expandafter\ifx\csname urlstyle\endcsname\relax
  \providecommand{\doi}[1]{doi: #1}\else
  \providecommand{\doi}{doi: \begingroup \urlstyle{rm}\Url}\fi

\bibitem[Abbasi-Yadkori et~al.(2011)Abbasi-Yadkori, P{\'a}l, and
  Szepesv{\'a}ri]{abbasi2011improved}
Yasin Abbasi-Yadkori, D{\'a}vid P{\'a}l, and Csaba Szepesv{\'a}ri.
\newblock Improved algorithms for linear stochastic bandits.
\newblock \emph{Advances in neural information processing systems}, 24, 2011.

\bibitem[Auer et~al.(2002)Auer, Cesa-Bianchi, and Fischer]{auer2002finite}
Peter Auer, Nicolo Cesa-Bianchi, and Paul Fischer.
\newblock Finite-time analysis of the multiarmed bandit problem.
\newblock \emph{Machine learning}, 47\penalty0 (2):\penalty0 235--256, 2002.

\bibitem[Bai et~al.(2021)Bai, Jin, Wang, and Xiong]{bai2021sample}
Yu~Bai, Chi Jin, Huan Wang, and Caiming Xiong.
\newblock Sample-efficient learning of {Stackelberg} equilibria in general-sum
  games.
\newblock \emph{Advances in Neural Information Processing Systems},
  34:\penalty0 25799--25811, 2021.

\bibitem[Conitzer and Sandholm(2006)]{conitzer2006computing}
Vincent Conitzer and Tuomas Sandholm.
\newblock Computing the optimal strategy to commit to.
\newblock In \emph{Proceedings of the 7th ACM Conference on Electronic
  Commerce}, pages 82--90, 2006.

\bibitem[Dong et~al.(2018)Dong, Roth, Schutzman, Waggoner, and
  Wu]{dong2018strategic}
Jinshuo Dong, Aaron Roth, Zachary Schutzman, Bo~Waggoner, and Zhiwei~Steven Wu.
\newblock Strategic classification from revealed preferences.
\newblock In \emph{Proceedings of the 2018 ACM Conference on Economics and
  Computation}, pages 55--70, 2018.

\bibitem[Dong et~al.(2021)Dong, Yang, and Ma]{dong2021provable}
Kefan Dong, Jiaqi Yang, and Tengyu Ma.
\newblock Provable model-based nonlinear bandit and reinforcement learning:
  Shelve optimism, embrace virtual curvature.
\newblock \emph{Advances in Neural Information Processing Systems},
  34:\penalty0 26168--26182, 2021.

\bibitem[Ferber and Weiss(1999)]{ferber1999multi}
Jacques Ferber and Gerhard Weiss.
\newblock \emph{Multi-agent systems: an introduction to distributed artificial
  intelligence}, volume~1.
\newblock Addison-wesley Reading, 1999.

\bibitem[Filar and Vrieze(2012)]{filar2012competitive}
Jerzy Filar and Koos Vrieze.
\newblock \emph{Competitive Markov decision processes}.
\newblock Springer Science \& Business Media, 2012.

\bibitem[Foster et~al.(2021)Foster, Kakade, Qian, and
  Rakhlin]{foster2021statistical}
Dylan~J Foster, Sham~M Kakade, Jian Qian, and Alexander Rakhlin.
\newblock The statistical complexity of interactive decision making.
\newblock \emph{arXiv preprint arXiv:2112.13487}, 2021.

\bibitem[Gerstgrasser and Parkes(2022)]{gerstgrasser2022oracles}
Matthias Gerstgrasser and David~C Parkes.
\newblock Oracles \& followers: Stackelberg equilibria in deep multi-agent
  reinforcement learning.
\newblock \emph{arXiv preprint arXiv:2210.11942}, 2022.

\bibitem[Goodrich et~al.(2008)Goodrich, Schultz, et~al.]{goodrich2008human}
Michael~A Goodrich, Alan~C Schultz, et~al.
\newblock Human--robot interaction: a survey.
\newblock \emph{Foundations and Trends{\textregistered} in Human--Computer
  Interaction}, 1\penalty0 (3):\penalty0 203--275, 2008.

\bibitem[Hardt et~al.(2016)Hardt, Megiddo, Papadimitriou, and
  Wootters]{hardt2016strategic}
Moritz Hardt, Nimrod Megiddo, Christos Papadimitriou, and Mary Wootters.
\newblock Strategic classification.
\newblock In \emph{Proceedings of the 2016 ACM conference on Innovations in
  Theoretical Computer Science}, pages 111--122, 2016.

\bibitem[Ho et~al.(2014)Ho, Slivkins, and Vaughan]{ho2014adaptive}
Chien-Ju Ho, Aleksandrs Slivkins, and Jennifer~Wortman Vaughan.
\newblock Adaptive contract design for crowdsourcing markets: Bandit algorithms
  for repeated principal-agent problems.
\newblock In \emph{Proceedings of the fifteenth ACM conference on Economics and
  computation}, pages 359--376, 2014.

\bibitem[Huang et~al.(2021)Huang, Huang, Kakade, Lee, Lei, Wang, and
  Yang]{huang2021optimal}
Baihe Huang, Kaixuan Huang, Sham Kakade, Jason~D Lee, Qi~Lei, Runzhe Wang, and
  Jiaqi Yang.
\newblock Optimal gradient-based algorithms for non-concave bandit
  optimization.
\newblock \emph{Advances in Neural Information Processing Systems},
  34:\penalty0 29101--29115, 2021.

\bibitem[Kao et~al.(2022)Kao, Wei, and Subramanian]{kao2022decentralized}
Hsu Kao, Chen-Yu Wei, and Vijay Subramanian.
\newblock Decentralized cooperative reinforcement learning with hierarchical
  information structure.
\newblock In \emph{International Conference on Algorithmic Learning Theory},
  pages 573--605. PMLR, 2022.

\bibitem[Kleinberg and Leighton(2003)]{kleinberg2003value}
Robert Kleinberg and Tom Leighton.
\newblock The value of knowing a demand curve: Bounds on regret for online
  posted-price auctions.
\newblock In \emph{44th Annual IEEE Symposium on Foundations of Computer
  Science, 2003. Proceedings.}, pages 594--605. IEEE, 2003.

\bibitem[Kober et~al.(2013)Kober, Bagnell, and Peters]{kober2013reinforcement}
Jens Kober, J~Andrew Bagnell, and Jan Peters.
\newblock Reinforcement learning in robotics: A survey.
\newblock \emph{The International Journal of Robotics Research}, 32\penalty0
  (11):\penalty0 1238--1274, 2013.

\bibitem[Langford and Zhang(2007)]{langford2007epoch}
John Langford and Tong Zhang.
\newblock The epoch-greedy algorithm for contextual multi-armed bandits.
\newblock \emph{Advances in neural information processing systems}, 20\penalty0
  (1):\penalty0 96--1, 2007.

\bibitem[Lauffer et~al.(2022)Lauffer, Ghasemi, Hashemi, Savas, and
  Topcu]{lauffer2022no}
Niklas Lauffer, Mahsa Ghasemi, Abolfazl Hashemi, Yagiz Savas, and Ufuk Topcu.
\newblock No-regret learning in dynamic {S}tackelberg games.
\newblock \emph{arXiv preprint arXiv:2202.04786}, 2022.

\bibitem[Lillicrap et~al.(2015)Lillicrap, Hunt, Pritzel, Heess, Erez, Tassa,
  Silver, and Wierstra]{lillicrap2015continuous}
Timothy~P Lillicrap, Jonathan~J Hunt, Alexander Pritzel, Nicolas Heess, Tom
  Erez, Yuval Tassa, David Silver, and Daan Wierstra.
\newblock Continuous control with deep reinforcement learning.
\newblock \emph{arXiv preprint arXiv:1509.02971}, 2015.

\bibitem[Liu and Chen(2016)]{liu2016bandit}
Yang Liu and Yiling Chen.
\newblock A bandit framework for strategic regression.
\newblock \emph{Advances in Neural Information Processing Systems}, 29, 2016.

\bibitem[Marecki et~al.(2012)Marecki, Tesauro, and Segal]{marecki2012playing}
Janusz Marecki, Gerry Tesauro, and Richard Segal.
\newblock Playing repeated {S}tackelberg games with unknown opponents.
\newblock In \emph{Proceedings of the 11th International Conference on
  Autonomous Agents and Multiagent Systems-Volume 2}, pages 821--828, 2012.

\bibitem[Rajaraman et~al.(2021)Rajaraman, Han, Yang, Liu, Jiao, and
  Ramchandran]{rajaraman2021value}
Nived Rajaraman, Yanjun Han, Lin Yang, Jingbo Liu, Jiantao Jiao, and Kannan
  Ramchandran.
\newblock On the value of interaction and function approximation in imitation
  learning.
\newblock \emph{Advances in Neural Information Processing Systems},
  34:\penalty0 1325--1336, 2021.

\bibitem[Russo and Van~Roy(2013)]{russo2013eluder}
Daniel Russo and Benjamin Van~Roy.
\newblock Eluder dimension and the sample complexity of optimistic exploration.
\newblock \emph{Advances in Neural Information Processing Systems}, 26, 2013.

\bibitem[Sallab et~al.(2017)Sallab, Abdou, Perot, and Yogamani]{sallab2017deep}
Ahmad~EL Sallab, Mohammed Abdou, Etienne Perot, and Senthil Yogamani.
\newblock Deep reinforcement learning framework for autonomous driving.
\newblock \emph{Electronic Imaging}, 2017\penalty0 (19):\penalty0 70--76, 2017.

\bibitem[Shalev-Shwartz et~al.(2016)Shalev-Shwartz, Shammah, and
  Shashua]{shalev2016safe}
Shai Shalev-Shwartz, Shaked Shammah, and Amnon Shashua.
\newblock Safe, multi-agent, reinforcement learning for autonomous driving.
\newblock \emph{arXiv preprint arXiv:1610.03295}, 2016.

\bibitem[Tambe(2011)]{tambe2011security}
Milind Tambe.
\newblock \emph{Security and Game Theory: Algorithms, Deployed Systems, Lessons
  Learned}.
\newblock Cambridge University Press, 2011.

\bibitem[von Stackelberg(2010)]{von2010market}
Heinrich von Stackelberg.
\newblock \emph{Market Structure and Equilibrium}.
\newblock Springer Science \& Business Media, 2010.

\bibitem[Wainwright(2019)]{wainwright2019high}
Martin~J Wainwright.
\newblock \emph{High-Dimensional Statistics: A Non-Asymptotic Viewpoint},
  volume~48.
\newblock Cambridge University Press, 2019.

\bibitem[Wang et~al.(2005)Wang, Kulkarni, and Poor]{wang2005bandit}
Chih-Chun Wang, Sanjeev~R Kulkarni, and H~Vincent Poor.
\newblock Bandit problems with side observations.
\newblock \emph{IEEE Transactions on Automatic Control}, 50\penalty0
  (3):\penalty0 338--355, 2005.

\bibitem[Wooldridge(2009)]{wooldridge2009introduction}
Michael Wooldridge.
\newblock \emph{An introduction to multiagent systems}.
\newblock John wiley \& sons, 2009.

\bibitem[Xie et~al.(2021)Xie, Losey, Tolsma, Finn, and Sadigh]{xie2021learning}
Annie Xie, Dylan Losey, Ryan Tolsma, Chelsea Finn, and Dorsa Sadigh.
\newblock Learning latent representations to influence multi-agent interaction.
\newblock In \emph{Conference on robot learning}, pages 575--588. PMLR, 2021.

\bibitem[Yang et~al.(2022)Yang, Zheng, Ratliff, Boots, and
  Smith]{yang2022stackelberg}
Boling Yang, Liyuan Zheng, Lillian~J Ratliff, Byron Boots, and Joshua~R Smith.
\newblock Stackelberg maddpg: Learning emergent behaviors via information
  asymmetry in competitive games.
\newblock 2022.

\bibitem[Yu et~al.(2022)Yu, Xu, and Chen]{yu2022learning}
Yaolong Yu, Haifeng Xu, and Haipeng Chen.
\newblock Learning correlated {S}tackelberg equilibrium in general-sum
  multi-leader-single-follower games.
\newblock \emph{arXiv preprint arXiv:2210.12470}, 2022.

\bibitem[Zhang et~al.(2021)Zhang, Yang, and Ba{\c{s}}ar]{zhang2021multi}
Kaiqing Zhang, Zhuoran Yang, and Tamer Ba{\c{s}}ar.
\newblock Multi-agent reinforcement learning: A selective overview of theories
  and algorithms.
\newblock \emph{Handbook of Reinforcement Learning and Control}, pages
  321--384, 2021.

\bibitem[Zhong et~al.(2021)Zhong, Yang, Wang, and Jordan]{zhong2021can}
Han Zhong, Zhuoran Yang, Zhaoran Wang, and Michael~I Jordan.
\newblock Can reinforcement learning find {S}tackelberg-{N}ash equilibria in
  general-sum {M}arkov games with myopic followers?
\newblock \emph{arXiv preprint arXiv:2112.13521}, 2021.

\bibitem[Zhu et~al.(2022)Zhu, Bates, Yang, Wang, Jiao, and
  Jordan]{zhu2022sample}
Banghua Zhu, Stephen Bates, Zhuoran Yang, Yixin Wang, Jiantao Jiao, and
  Michael~I Jordan.
\newblock The sample complexity of online contract design.
\newblock \emph{arXiv preprint arXiv:2211.05732}, 2022.

\end{thebibliography}
\bibliographystyle{plainnat}

\newpage
\appendix
\onecolumn

\section{Proofs in Section~\ref{Sec_basic_upper_lower_bounds}}\label{Append_proofs_1}

\subsection{Proof of Theorem~\ref{Thm_exp_lower_bound}}\label{proof:Thm_exp_lower_bound}
\begin{proof}
Consider Example~\ref{ex:relu_bandit}.
The expected reward is given by
\begin{equation}
    h_\theta(a,b) := \theta \cdot \phi(a,b) = (1-b) \theta_{-d}\cdot a + b (1-\Delta),
\end{equation}
Optimizing over $b\in[0,1]$ yields
\begin{equation}
    \hupper_\theta(a) = \max\{1-\Delta, \theta_{-d}\cdot a\}.
\end{equation}
Note that for any $a\in\cA$ such that $\theta_{-d}\cdot a < 1-\Delta$, the best response of the follower is $b=1$, yielding an expected reward of $1-\Delta$; for any $a\in\cA$ such that $\theta_{-d}\cdot a \geq  1-\Delta$, the best response of the follower is $b=0$, yielding an expected reward of $\theta_{-d} \cdot a$. The optimal joint response $a = \theta_{-d}$ and  $b=0$ achieves the optimal expected reward of $\|\theta_{-d}\| = 1 > 1-\Delta$. 
From the leader's perspective, this now reduces to the problem of a ReLU bandit considered in~\citet{dong2021provable}, since the response provides no information until the average regret falls below $\Delta$.\footnote{Same as in \citet{dong2021provable}, we allow the reward and response observations to be noiseless. We believe, however, the proof of \citet[Theorem~5.1]{dong2021provable} has a small gap, where the packing number should be computed for radius $\sqrt{\eps}$ instead of $\eps$. This lower bound can be further improved if we assume noisy observations.} Thus we have
\begin{align*}
    \inf_{\hat \pi} \sup_{\theta\in\Theta}\cR(T) \geq \Omega(T^{1-\frac{2}{d-2}}).
\end{align*}
\end{proof}

\subsection{Proof of Theorem~\ref{Thm_general_exp_upper_bound}}\label{proof:Thm_general_exp_upper_bound}
\begin{proof}
    Let $\mathcal{H}(\epsilon)$ be a minimal $\epsilon$-covering of $\mathcal{H}$ under the metric $\|\cdot\|_\infty$. Let $$\mathcal{A}(\epsilon) = \big\{\argmax_{a\in\mathcal{A}} \max_{b\in{\mathcal{B}}} h(a, b) \mid  h\in\mathcal{H}(\epsilon)\big\},$$ where we break ties arbitrarily when the optimal action is non-unique.
Note that we have $|\mathcal{A}(\epsilon)| \leq |\mathcal{H}(\epsilon)|\leq N(\epsilon)$. Let $h^\star$  be the true reward function. By the definition of a covering, there exists some $ h_\epsilon\in\mathcal{H}(\epsilon)$ such that $\|h^\star -  h_\epsilon\|_\infty\leq \epsilon$. Thus we have
\begin{align*}
 \cR(T) = \sum_{t=1}^T \mathbb{E}[\hupper^\star(a^*) - \hupper^\star(a_t)] \leq \epsilon T + \sum_{t=1}^T \mathbb{E}[\hupper_\epsilon^\star(a^*) - \hupper^\star_\epsilon(a_t)]. 
\end{align*}
We know that the optimal action for $\hupper_\epsilon$ must be inside the set $\mathcal{A}(\epsilon)$. Thus  any worst-case optimal no-regret algorithm on the set $\mathcal{A}(\epsilon)$ gives a regret of $\sqrt{|\mathcal{A}(\epsilon)|T}\leq  \sqrt{N(\epsilon) T}$. This gives that
\begin{align*}
 \cR(T) \leq \epsilon T + \sqrt{N(\epsilon)T}. 
\end{align*}
Taking infimum over $\epsilon$ finishes the proof. 
\end{proof}

\section{Proofs in Section~\ref{Sec_ucb}}

\subsection{Proof of Lemma~\ref{Lemma_choice_of_alpha}}
\begin{proof}
    Recall the notation from Example~\ref{Ex_theta_aplusb}: let $\theta_t^{(b)} = \Pi_\cA(\hat\theta_t)$ for $t \ge 2$, with $\hat\theta_t := \frac{1}{t-1}\sum_{i=1}^{t-1} \hat{b}_i$. The first round incurs at most a constant regret and can be ignored. It suffices to show that, with probability at least $1-\delta$,
    \begin{equation}\label{Eqn_proof_lemma_choice_alpha_goal}
        \|\theta-\theta_t^{(b)}\| \le \frac{\alpha_t}{\sqrt{t}}
    \end{equation}
    for $\alpha_t = \Theta\big(\sigma_b \sqrt{d+\log\frac{T}{\delta}}\big)$.

    First, we bound the distance between $\hat\theta_t$ and $\theta$. By our assumption,
    \begin{equation*}
        \|\hat\theta_t - \theta\| = \Big\|\frac{1}{t-1}\sum_{i=1}^{t-1}w_i\Big\|,
    \end{equation*}
    where $w_1,\ldots,w_t$ are i.i.d. zero-mean $\sigma_b$-sub-Gaussian. We proceed using a covering argument. Construct $U\subseteq \S^{d-1}$ such that
    \begin{equation}
        \inf_{v\in\S^{d-1}} \sup_{u\in U} u\cdot v \ge \frac{1}{2}.
    \end{equation}
    Note that $\|u-v\| = \sqrt{2-2u\cdot v}$ for $u,v\in\S^{d-1}$. Hence, equivalently, we may choose $U$ as a minimal $1$-covering of $\S^{d-1}$ in Euclidean metric. Then 
    \begin{equation}
        \log |U| \le \log N^\text{int}(\S^{d-1},1,\|\cdot\|) \le \log M(\B^d, 1, \|\cdot\|) = \Theta(d),
    \end{equation}
    where $N^\text{int}$ and $M$ denote the internal covering number and the packing number of the space under a given metric. The choice of $U$ ensures that
    \begin{equation}\label{Eqn_proof_theta_aplusb_covering_bound}
        \|w\| \le 2\sup_{u\in U} u\cdot w
    \end{equation}
    for all $w \in\R^d$, and ignoring the constant factor, we may focus on upper bounding $\sup_{u\in U} \sum_{i=1}^{t-1} u\cdot w_i$.
    
    For each choice of $u\in U$, let $Z_{u,i} = u\cdot w_i$, so that $Z_{u,1},\ldots,Z_{u,t-1}$ are i.i.d. zero-mean $\sigma_b$-sub-Gaussian by definition of sub-Gaussian random vectors. By Hoeffding's inequality for sub-Gaussian random variables, we have
    \begin{equation}
        \Pp\bigg(\sum_{i=1}^t Z_{u,i} > x \bigg) \le \exp\Big(-\frac{x^2}{2t\sigma_b^2}\Big)
    \end{equation}
    for all $x > 0$.
    Applying union bound over $U$ and using \eqref{Eqn_proof_theta_aplusb_covering_bound} gives
    \begin{equation}
        \Pp\left(\bigg\|\sum_{i=1}^t w_i\bigg\| \ge 2x \right) \le \Pp\left(\sup_{u\in U} \sum_{i=1}^t Z_{u,i} \ge x \right) \le |U|\exp\Big(-\frac{x^2}{2t\sigma_b^2}\Big).
    \end{equation}
    Choosing $x = \sigma_b\sqrt{2t\log(|U|T)} \lesssim \sigma_b \sqrt{t (d+\log \frac{T}{\delta})}$ ensures that, by another union bound over $t\in[T]$,
    \begin{equation}
        \|\hat\theta_t-\theta\| \lesssim \sigma_b \sqrt{t^{-1}\big(d+\log \frac{T}{\delta}\big)}
    \end{equation}
    with probability at least $1-\delta$. By the triangle inequality and the definition of projection,
    \begin{equation}
        \|\theta_t^{(b)}-\theta\| \le \|\theta_t^{(b)}-\hat\theta_t\| + \|\hat\theta_t-\theta\| \le 2 \|\hat\theta_t-\theta\| \lesssim \sigma_b \sqrt{t^{-1}\big(d+\log \frac{T}{\delta}\big)}
    \end{equation}
    with the same probability. This gives \eqref{Eqn_proof_lemma_choice_alpha_goal} and completes the proof.
\end{proof}

\subsection{Proof of Proposition~\ref{Prop_theta_a_b_upper_bound}}
\begin{proof}
    We will condition upon the validity of the confidence sets, which happens with probability at least $1-\delta$ per our choice of $\{\alpha_t\}_{t\in[T]}$.

    UCB always chooses $a_t$ in the confidence set $\Theta_t$, with radius of order $\bO\big(\sigma_b\sqrt{t^{-1} (d+\log \frac{T}{\delta})}\big)$. When $\theta^\star\in\Theta_t$, we have $\|a_t - \theta^\star\| \lesssim \sigma_b\sqrt{t^{-1} (d+\log \frac{T}{\delta})}$. Since both $a_t$ and $\theta^\star$ are unit vectors, we have
    \begin{align*}
        \cR_{UCB}(T) &\leq  2\delta T + \sum_{t=1}^T \big(1 - \theta^\star \cdot a_t\big) = 2\delta T + 2 + \frac{1}{2} \sum_{t=1}^T \|\theta^\star - a_t\|^2 \\
        &\lesssim 2\delta T + \sum_{t=2}^T \frac{\sigma_b^2}{t}\Big(d+\log \frac{T}{\delta}\Big) = \bO\bigg(\delta T + \sigma_b^2 \log T\cdot \Big(d + \log \frac{T}{\delta}\Big)\bigg),
    \end{align*}
    where the term $2\delta T$ bounds the contribution of the event that the confidence sets fails to be all valid. Choosing $\delta = 1/T$ gives our desired bound.
\end{proof}

\subsection{Proof of Proposition~\ref{Prop_guide_explore_upper_bound_strong_reduction}}
\begin{proof}
    After the first round, the leader's task reduces to a linear bandit with action space $\cA_1$: only actions within $\cA_1$ will be played, and the reward is linear in this region.
    As is well known for linear bandit (e.g., \cite{russo2013eluder}), with probability $1-\delta$, the regret in this linear stage (i.e., excluding the first round) is upper bounded by
    \begin{equation*}
        2\delta T + \bO\big(\sqrt{d\log T \cdot (d \log T + \log\delta^{-1}) \cdot T}\big).
    \end{equation*}
    The first round adds at most a constant to this and can be ignored. By choosing $\delta = T^{-1}$, we have
    \begin{equation}
        \cR_{UCB}(T) \le \widetilde{\bO}(d\sqrt{T}).
    \end{equation}
\end{proof}

\subsection{Proof of Proposition~\ref{Prop_guide_explore_upper_bound_weak_reduction}}
\begin{proof}
    Let $\Theta_1 = \{\theta_a\in\S^{d-1}|\theta_a\cdot b_1\ge \zeta\}\times\{b_1\}$, and denote the true parameter by $\theta^\star = (\theta_a^\star,\theta_b^\star)$. By our assumption on the problem structure, we have $\theta^\star_a \in \Theta^{(b)}$.
    
    As in the proof of Theorem~\ref{Thm_general_exp_upper_bound}, let $\Theta(\eps)$ be a minimal $\eps$-covering of $\Theta_1$ in Euclidean metric, with $\eps>0$ to be specified later. In particular, there is some $\tilde{\theta}_a \in \Theta_1$ with $\|\tilde{\theta}_a - \theta_a^\star\| \le \eps$. Let $\cA(\eps) = \{ \argmax_{a\in\cA} \ReLU(\theta_a \cdot a - \Delta) \mid \theta_a \in \Theta(\eps)\}$, where we break tie arbitrarily when the optimal action is non-unique. Note that $|\cA(\eps)| \le |\Theta(\eps)| = N(\Theta_1,\eps,\|\cdot\|)$.

    Now, let the leader play UCB on the discrete action set $\cA(\eps)$ after the first round. The regret satisfies
    \begin{equation}\label{Eqn_proof_prop_guide_explore_weak_reduction}
        \cR(T) \le 1 + \sum_{t=2}^T \mathbb{E}\big[\hupper^\star(a^*) - \hupper^\star(a_t) \big] \le 1 + T\cdot \mathbb{E}\big[\hupper^\star(a^*) - \hupper^\star(\tilde{a}^*)\big] + \sum_{t=1}^T  \mathbb{E}\big[\hupper^\star(\tilde{a}^*) - \hupper^\star(a_t)\big],
    \end{equation}
    where $a^* = \theta_a^\star$ and $\tilde{a}^* \in \argmax_{a\in\cA(\eps)} \hupper^\star(a)$. Note that $\hupper^\star(\tilde{a}^*) \ge \hupper^\star(\tilde{\theta}_a) \ge \hupper^\star(a^*) - \eps$ by our choice of $\tilde{\theta}_a$ and $\cA(\eps)$, the second term in \eqref{Eqn_proof_prop_guide_explore_weak_reduction} is at most $\eps T$. The third term, the regret of UCB on $\cA(\eps)$, is bounded by $\bO(\sqrt{N(\Theta_1,\eps,\|\cdot\|)\cdot T})$ in expectation.

    It remains to bound $N(\Theta_1,\eps,\|\cdot\|)$. Note that for any $\theta_a,\theta_a'\in\Theta_1$, we have
    \begin{align*}
        \theta_a \cdot \theta_a' &= (\theta_a\cdot b_1)(\theta_a'\cdot b_1) + (\theta_a - (\theta_a\cdot b_1) b_1)\cdot (\theta_a' - (\theta_a'\cdot b_1) b_1) \\
        &\ge \zeta^2 - \|\theta_a - (\theta_a\cdot b_1) b_1\| \|\theta_a' - (\theta_a'\cdot b_1) b_1\| \\
        &\ge \zeta^2 - (1 - \zeta^2) = 2\zeta^2 - 1.
    \end{align*}
    Equivalently, $\|\theta_a - \theta_a'\| = \sqrt{2-2\theta_a\cdot\theta_a'} \le 2\sqrt{1-\zeta^2} = 2C_\zeta$. Thus, the covering number of $\Theta_1$ is upper bounded by $\big(\frac{KC_\zeta}{\eps}^d)$ for some absolute constant $K$, which yields a regret bound of $1 + \eps T + \bO(\sqrt{K^d C_\zeta^d T / \eps^d})$. Choosing $\eps \asymp (KC_\zeta)^{\frac{d}{d+2}} T^{-\frac{1}{d+2}}$ reduces this upper bound to $\bO\Big(C_\zeta^{\frac{d}{d+2}} T^{\frac{d+1}{d+2}}\Big)$ as desired.
\end{proof}

\section{Proofs in Section~\ref{Sec_beyond_ucb}}

\subsection{Proof of Proposition~\ref{Prop_nonlinear_bandit_upper}}
\begin{proof}
    Let the leader run the phased elimination algorithm \citet[Algorithm~6]{huang2021optimal} using the response $b^*_\theta(a_t)$ as the proxy reward to maximize. This proxy reward, in expectation, is a homogeneous polynomial of degree $2k-1$. By Corollary~3.16 in \citet{huang2021optimal}, the algorithm achieves
    \begin{equation}
        \widehat{\cR}(T) \le \widetilde{\bO}\big(\sqrt{d^{2k-1} T}\big),
    \end{equation}
    where $\widehat{\cR}(T) = \sum_{t=1}^T 1-b^*_\theta(a_t)$ is the proxy regret measured based on the the proxy reward (i.e., absolute response). Note that the reward is maximized exactly when the proxy reward is maximized. Thus, the Lipschitz property \eqref{Eqn_ex_nonlinear_lipschitz} suggests that
    \begin{equation}
        \cR(T) \le \frac{2k}{2k-1} \widehat{\cR}(T)\le \widetilde{\bO}(\sqrt{d^{2k-1} T}).
    \end{equation}
\end{proof}

\end{document}